\documentclass[10pt, a4paper]{article}
\usepackage{lrec}
\usepackage{multibib}
\usepackage{graphicx}
\usepackage{tabularx}
\usepackage{soul}
\usepackage{booktabs}
\usepackage{subfig}
\usepackage{tabulary}

\usepackage{epstopdf}
\usepackage[latin1]{inputenc}

\usepackage{hyperref}
\usepackage{xstring}

\newcommand{\secref}[1]{\StrSubstitute{\getrefnumber{#1}}{.}{ }}

\title{A Corpus for Modeling Word Importance in Spoken Dialogue Transcripts}

\name{Sushant Kafle, Matt Huenerfauth}

\address{Rochester Institute of Technology\\
         1 Lomb Memorial Drive, Rochester, NY \\
         sushant@mail.rit.edu, matt.huenerfauth@rit.edu\\}

\abstract{
Motivated by a project to create a system for people who are deaf or hard-of-hearing that would use automatic speech recognition (ASR) to produce real-time text captions of spoken English during in-person meetings with hearing individuals, we have augmented a transcript of the Switchboard conversational dialogue corpus with an overlay of word-importance annotations, with a numeric score for each word, to indicate its importance to the meaning of each dialogue turn. Further, we demonstrate the utility of this corpus by training an automatic word importance labeling model; our best performing model has an F-score of $0.60$ in an ordinal 6-class word-importance classification task with an agreement (concordance correlation coefficient) of $0.839$ with the human annotators (agreement score between annotators is $0.89$). Finally, we discuss our intended future applications of this resource, particularly for the task of evaluating ASR performance, i.e. creating metrics that predict ASR-output caption text usability for DHH users better than Word Error Rate (WER).\\ \newline }

\begin{document}

\maketitleabstract

\section{Introduction}
There has been increasing interest among researchers of speech and language technology applications to identify the importance of individual words, for the overall meaning of the text. Depending on the context of how the importance of a word is defined, this task has found use in varieties of applications such as text summarization \cite{hong2014improving,yih2007multi}, text classification \cite{sheikh2016learning}, or speech synthesis \cite{mishra2007word}. 

Our laboratory is currently designing a system to benefit people who are deaf or hard-of-hearing (DHH) who are engaged in a live meeting with hearing colleagues. In many settings, sign language interpreting or professional captioning (where a human types the speech, displayed as text on a screen for the user), are unavailable, e.g. in impromptu conversations in the workplace.  A system that uses automatic speech recognition (ASR) to generate captions in real-time could display this text on mobile devices for DHH users, but text output from ASR systems inevitably contains errors. Thus, we were motivated to understand which words in the text were most important to the overall meaning, to inform our evaluation of ASR accuracy for this task. 

In this paper, we present a word-importance annotation of transcripts of the Switchboard corpus \cite{godfrey1992switchboard}. While our overall goal is to produce measures of ASR accuracy for our caption application; to demonstrate the use of this corpus, in this paper, we present models that predict word-importance in spoken dialogue transcripts.

\subsection{ASR Evaluation}
ASR researchers generally report the performance of their systems using a metric called Word Error Rate (WER). The metric considers the number of errors in the output of the ASR system, normalized by the number of words human actually said in the audio recording. While WER has been the most commonly used intrinsic measure for the evaluation of ASR, there have been criticisms of WER \cite{mccowan2004use,morris2004and}, and several researchers have recommended alternative measures to better predict human task-performance in applications that depend on ASR \cite{garofolo2000trec,mishra2011predicting,kafle2015}. 

Among these newly proposed metrics, a common theme has been: rather than simply counting the number of errors, it would be better to consider the importance of the individual words that are incorrect - suggesting that it would be better to more heavily penalize systems that make errors on words that are important (with the definition of importance based on the specific application or task). This approach of penalizing errors differentially has been shown to be useful in various application settings, e.g. in our research for DHH users, we have found that an evaluation metric designed for predicting the usability of an ASR-generated transcription as a caption text for these users could benefit from word importance information \cite{kafle2016}. However, estimating the importance of a word has been challenging for our team thus far, because we have lacked corpora of conversational dialogue with word-importance annotation, for training a word-importance model.

\subsection{Word Importance Estimation}
Prior research on identifying and scoring important words in a text has largely focused on the task of keyword extraction, which involves identifying a set of descriptive words in a document that serves as a dense summary of the document. Several automatic keyword extraction techniques have been investigated over the years, including unsupervised methods using, e.g. Term Frequency x Inverse Document Frequency (TF-IDF) weighting \cite{hacohen2005automatic}, word co-occurrence probability estimation \cite{matsuo2004keyword} -- as well as supervised methods that leverage various linguistic features from text to achieve strong predictive performance \cite{liu2011,liu2004text,hulth2003improved,sheeba2012improved}.

While this conceptualization of word importance as a keyword-extraction problem has led to positive results in the field of text summarization \cite{Litvak2008,wan2007towards,hong2014improving}, this approach may not generalize to other applications. For instance, given the sometimes meandering nature of topic transition in spontaneous speech dialogue \cite{sheeba2012improved}, applications that process transcripts of such dialogue may benefit from a model of word importance that is more local, i.e. based on the importance of a word at sentential, utterance, or local dialogue level, rather than at a document-level. Furthermore, the dyadic nature of dialogue, with interleaved contributions from multiple speakers, may require special consideration when evaluating word importance. In this paper, we present a corpus with annotation of word importance that could be used to support research into these complex issues.

\section{Defining Word Importance}
In eye-tracking studies of reading behavior, researchers have found that readers rarely glance at every word in a text sequentially: Instead, they sometimes regress (glance back at previous words), re-fixate on a word, or skip words entirely \cite{rayner1998eye}. This research supports the premise that some words are of higher importance than others, for readers. Analyses of eye-tracking recordings have revealed a relationship between these eye-movement behaviors and various linguistic features, e.g. word length or word predictability. In general, readers' gaze often skips over words that are shorter or more predictable \cite{rayner2011eye}. 

While eye-tracking suggests some features that may relate to readers' judgments of word importance, at least as expressed through their choice of eye fixations, we needed to develop a specific definition of word importance in order to develop annotation guidelines for our study.  Rather than ask annotators to consider specific features, e.g. word length, which may pre-suppose a particular model, we instead took a functional perspective, with our application domain in mind.  That is, we define word importance for spontaneous spoken conversation as the degree to which a reader of a transcript of the dialogue would be unable to understand the overall meaning of a conversational utterance (a single turn of dialogue) if that word had been ``dropped'' or omitted from the transcript.  This definition underlies our annotation scheme (in \secref{section-scheme:ref}) and suits our target application, i.e. evaluating ASR for real-time captioning of meetings. 

\begin{figure}[!ht]
\begin{center}
\includegraphics[width=0.45\textwidth]{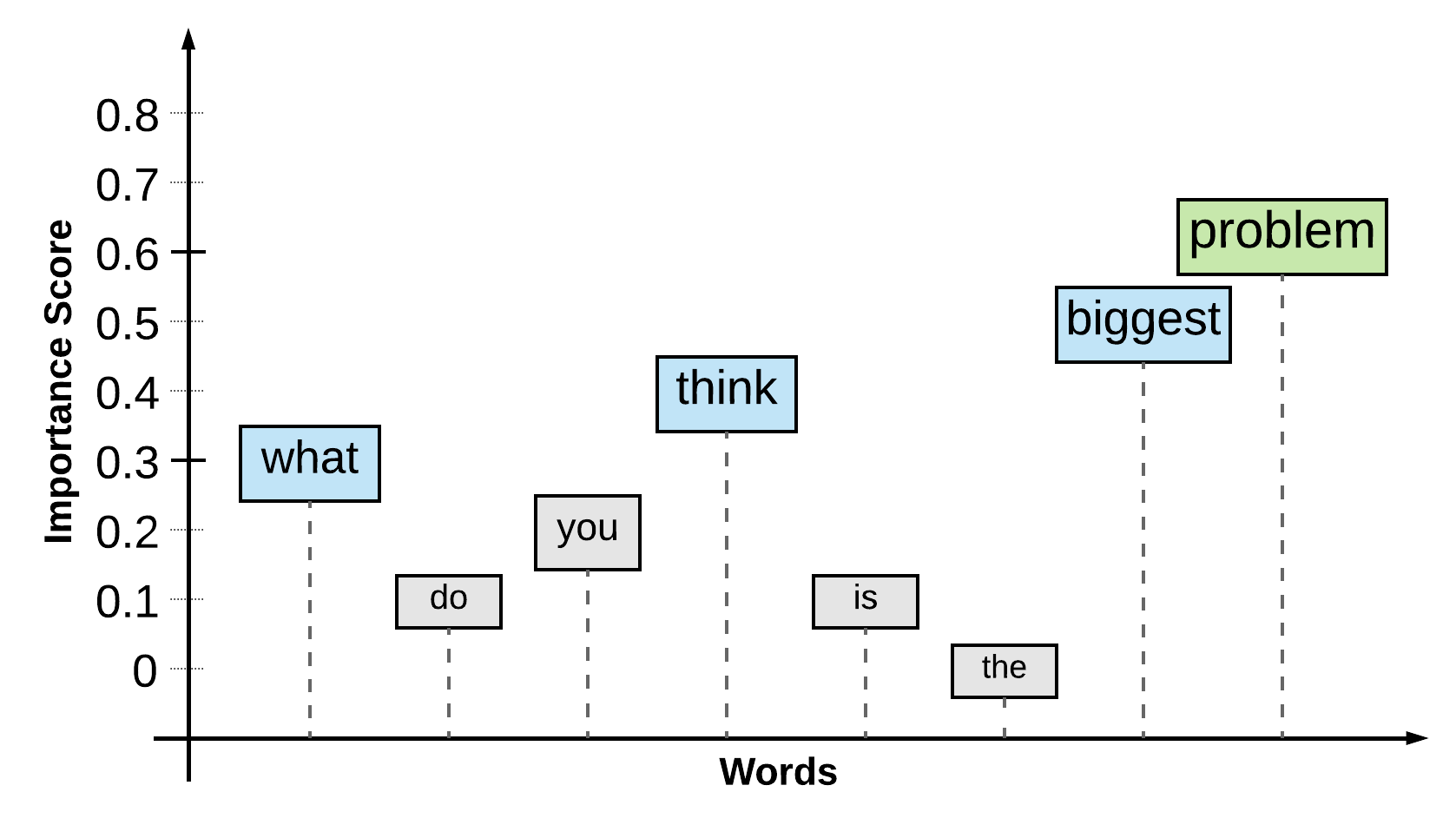} 
\caption{Visualization of importance scores assigned to words in a sentence by a human annotator on our project, with the height and font-size of words indicating their importance score (and redundant color coding: green for high-importance words with score above 0.6, blue for words with score between 0.3 and 0.6, and gray otherwise).}
\label{word-imp-example}
\end{center}
\end{figure}

In addition, for our annotation project, we defined word-importance as a single-dimensional property, which could be expressed on a continuous scale from 0.0 (not important at all to the meaning of the utterance) to 1.0 (very important). Figure \ref{word-imp-example} illustrates how numerical importance scores can be assigned to words in a sentence -- in fact, this figure displays actual scores assigned by a human annotator working on our project.  Of course, asking human annotators to assign specific numerical scores to quantify the importance of a word is not straightforward. In later sections, we discuss how we attempt to overcome the subjective nature of this task, to promote consistency between annotators, as we developed this annotated resource (see Section \secref{section-scheme:ref}). Section \secref{section-agreement} characterizes the level of agreement between our annotators on this task.

\section{Corpus Annotation}
\label{section-corpus-annotation}
The Switchboard corpus consists of audio recordings of approximately 260 hours of speech consisting of about 2,400 two-sided telephone conversations among 543 speakers (302 male, 241 female) from across the United States \cite{godfrey1992switchboard}. In January 2003, the Institute for Signal and Information Processing (ISIP) released written transcripts for the entire corpus, which consists of nearly 400,000 conversational turns. The ISIP transcripts include a complete lexicon list and automatic word alignment timing corresponding to the original audio files\footnote{https://www.isip.piconepress.com/projects/switchboard/}.

In our project, a pair of annotators have assigned word-importance scores to these transcripts. As of September 2017, they have annotated over 25,000 tokens, with the overlap of approximately 3,100 tokens.  With this paper, we announce the release\footnote{http://latlab.ist.rit.edu/lrec2018} of these annotations as a set of supplementary files, aligned to the ISIP transcripts.  Our annotation work continues, and we aim to annotate all of the Switchboard corpus and with a larger group of annotators.

\subsection{Annotation Scheme}
\label{section-scheme:ref}
To reduce the cognitive load on annotators and to promote consistency, we created the following annotation scheme:

\textbf{Range and Constraints.} Each word is assigned a numeric score between [0, 1], where 1 indicates a high importance score; the numeric score has the precision of 0.05. Importance scores are not meant to indicate an absolute proportion of the utterance's meaning represented by each word, i.e. the scores do not have to sum to 1. 

\textbf{Methodology.} Given an utterance (a speaker's single turn in the conversation), the annotator first considers the overall meaning conveyed by the utterance, with the help of the previous conversation history (if available). The annotator then scores each word based on its (direct or indirect) contribution to the utterance's meaning, using the rubric described in the Interpretation and Scoring section below.

\begin{table}[!ht]
\centering
\begin{tabular}{|l|l|}
\hline
\multicolumn{1}{|c|}{\textbf{Range}} & \multicolumn{1}{c|}{\textbf{Description}}                                                                                                                                     \\ \hline
{[}0 - 0.3)                          & \begin{tabular}[c]{@{}l@{}}Words that are of least importance - these \\ words can be easily omitted from the text \\ without much consequence.\end{tabular}                         \\ \hline
{[}0.3 - 0.6)                        & \begin{tabular}[c]{@{}l@{}}Words that are fairly important - omitting\\ these words will take away some important \\details from the utterance.\end{tabular} \\ \hline
{[}0.6 - 1{]}                        & \begin{tabular}[c]{@{}l@{}}Words that are of high importance - omitting \\ these words will change the message \\of the utterance quite significantly.\end{tabular}                \\ \hline
\end{tabular}
\caption{Guidance for the annotators to promote consistency and uniformity in the use of numerical scores.}
\label{rating-scheme}
\end{table}

\textbf{Rating Scheme}. To help annotators calibrate their scores, Table \ref{rating-scheme} provides some recommendations for how to select word-importance scores in various numerical ranges.

\textbf{Interpretation and Scoring.} Annotators should consider how their understanding of the utterance would be affected if this word had been ``dropped,'' i.e. replaced with a blank space (``\underline{\hspace{1cm}}''). Since these are conversations between pairs of speakers, annotators should consider how much the other person in the conversation would have difficulty understanding the speaker's message if that word had been omitted, i.e. if they had not heard that word intelligibly.

\section{Inter-annotator Agreement}
\label{section-agreement}
There were 3,100 tokens in our ``overlap'' set, i.e. the subset of transcripts independently labeled by both annotators.  This set was used as the basis for calculating inter-annotator agreement. Since scores were nearly continuous (ranges [0,1] with a precision of 0.05), we computed the Concordance Correlation Coefficient ($\rho_c$), also known as Lin's concordance correlation coefficient, as our primary metric for measuring the agreement between the annotators. This metric indicates how well a new test or measurement (X) reproduces a gold standard or measure (Y). Considering the annotations from one annotator as a gold standard, we can generalize this measure to compute the agreement between two annotators. Like other correlation coefficients, $\rho_c$ also ranges from -1 to 1; 1 being the score of perfect agreement. 

Concordance between the two measures can be characterized by the expected value of their squared difference as:

\begin{equation}
E[(Y - X)^2] = (\mu_y - \mu_x)^2 + \sigma^2_x + \sigma^2_y - 2\rho\sigma_x\sigma_y 
\end{equation}

where, $\rho$ is the correlation coefficient, $\mu_{x}$ and $\mu_{y}$ are the means of the population of the variables $X$ and $Y$, and $\sigma_{x}$ and $\sigma_{y}$ are their standard deviation. The expectation score coefficient (between -1 and 1) is calculated as follows:

\begin{equation}
\rho_c = \frac{2\rho S_x S_y}{(\bar{Y} - \bar{X})^2 + S^2_x + S^2_y}
\end{equation}

where, $\rho_c$ is the correlation coefficient, $\bar{X}$ and $\bar{Y}$ are the mean of $X$ and $Y$, and $S_x$ and $S_y$ are standard deviations.

\begin{figure}[!ht]
\begin{center}
\includegraphics[width=0.45\textwidth]{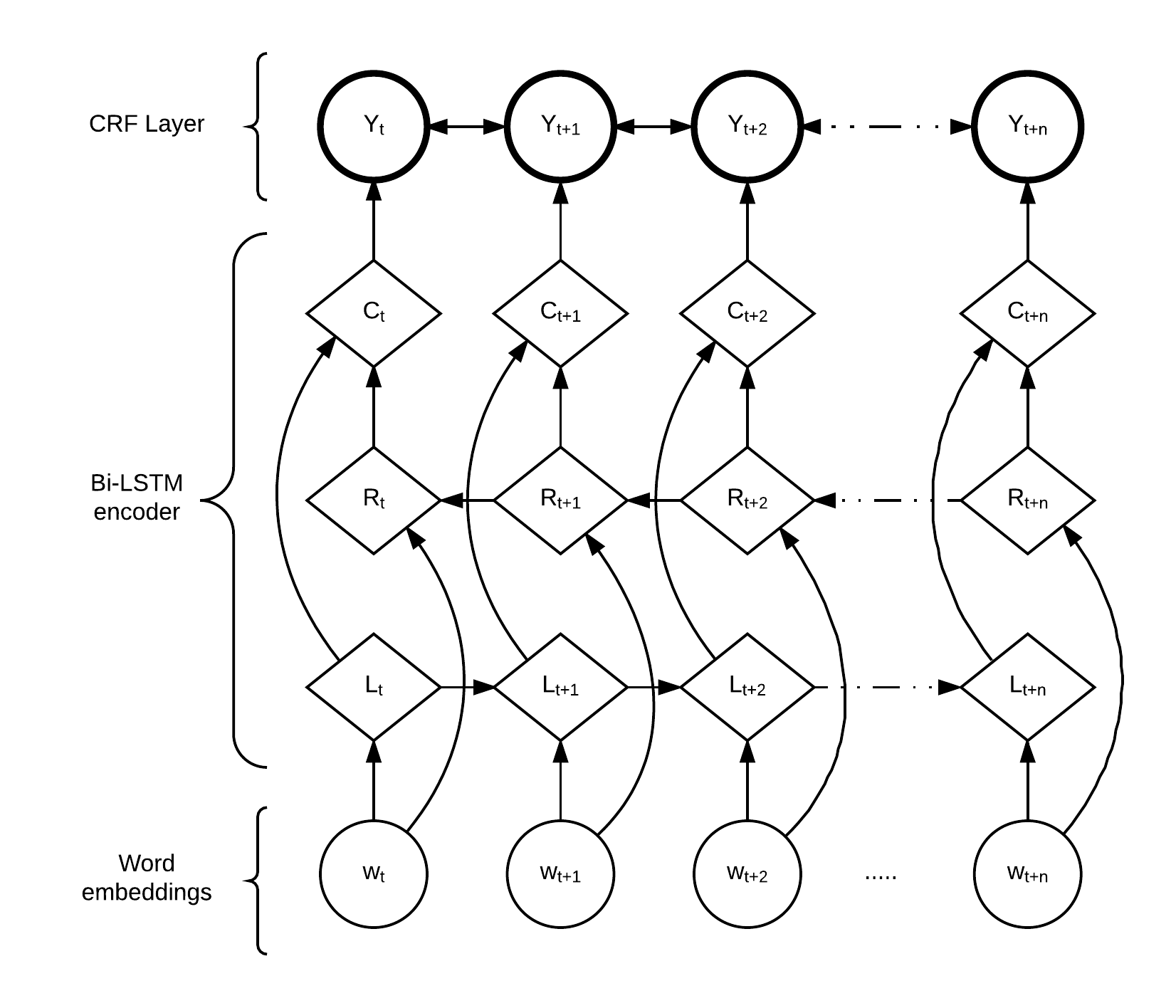} 
\caption={[General unfolded network structure of our model, adapted from \cite{LampleBSKD16}. The bottom layer represents word-embedding inputs, passed to bi-directional LSTM layers above. Each LSTM takes as input the hidden state from the previous time step and word embeddings from the current step, and outputs a new hidden state. $C_i$ concatenates hidden representations from LSTMs ($L_i$ and $R_i$) to represent the word at time $i$ in its context.]}
\label{neural-arch}
\end{center}
\end{figure}

We obtained an agreement score ($\rho_c$) of $0.89$ between our annotators, which we interpret as an acceptable level of agreement, given the subjective nature of the task of quantifying word importance in spoken dialogue transcripts.

\begin{figure*}[t!]
    \centering
    \subfloat[Normalized confusion matrix for LSTM-CRF]
    {{\includegraphics[width=0.4\textwidth]{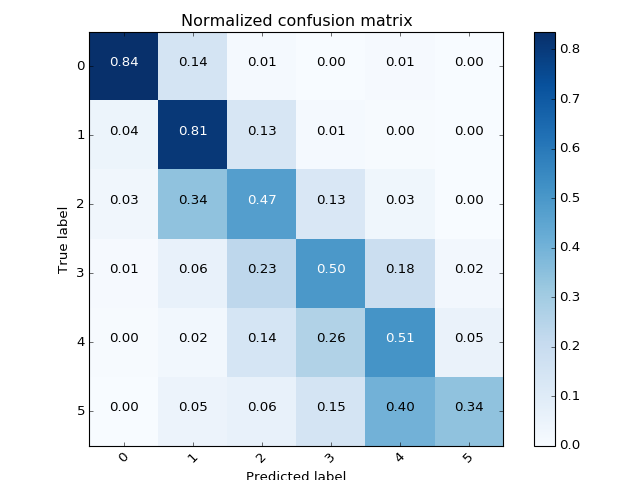}}}%
    \qquad
    \subfloat[Normalized confusion matrix for LSTM-SIG]
    {{\includegraphics[width=0.4\textwidth]{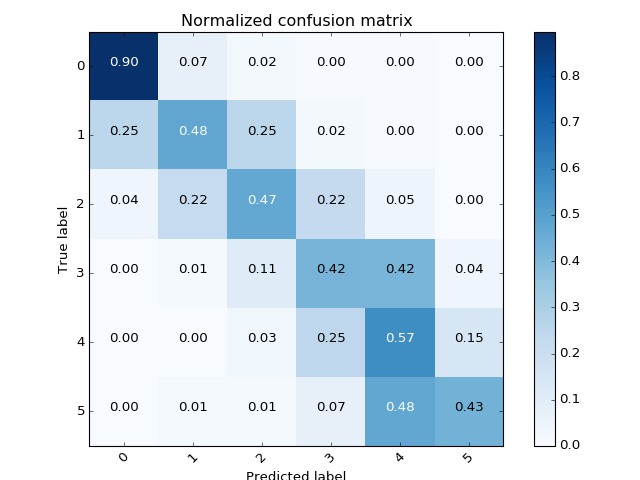}}}%
    \caption{Confusion matrices for each model for classification into 6 classes: $c_1$ = [0, 0.1), $c_2$ = [0.1, 0.3), and so forth.}%
    \label{conf-mat}%
\end{figure*}

\section{Automatic Prediction}
\label{automatic-prediction}
To demonstrate the use of this corpus, we trained a prediction model, by adopting the neural architecture described in \cite{LampleBSKD16} consisting of bidirectional LSTM encoders with a sequential Conditional Random Field (CRF) layer on top. Our input word tokens were first mapped to a sequence of pre-trained distributed embeddings \cite{pennington2014glove} and then combined with the learned character-based word representations to get the final word representation. As shown in Figure \ref{neural-arch}, the bidirectional LSTM encoders are used to create a context-aware representation of each word. The hidden representations from each LSTM were concatenated to obtain a final representation, conditioned on the whole sentence. The CRF layer uses this representation to look for the most optimal state ($Y$) sequence through all the possible state configurations.

The neural framework was implemented using Tensorflow, and the code is publicly available\footnote{https://github.com/SushantKafle/speechtext-wimp-labeler}. The word embeddings were initialized with publicly available pre-trained glove vectors \cite{pennington2014glove}. The embeddings for characters were set to length 100 and were initialized randomly. The LSTM layer size was set to 300 in each direction for word- and 100 for character-level components. Parameters were optimized using the Adam \cite{kingma2014adam} optimizer, with the learning rate initialized at 0.001 with a decay rate of 0.9, and sentences were grouped into batches of size 20. We applied a dropout with a probability of 0.5 during training on word embeddings.

We investigated two variations of this model: (i) a bidirectional LSTM model with sequential CRF layer on top (LSTM-CRF) treating the problem as a discrete classification task, (ii) a new bidirectional LSTM model with a sigmoid layer on top (LSTM-SIG) for a continuous prediction. The LSTM-CRF models the prediction task as a classification problem, using a fixed number of non-ordinal class labels. In contrast, the LSTM-SIG model provides a continuous prediction, using a sigmoid nonlinearity to bound the prediction scores between 0 and 1. Using a square loss, we train this model to directly learn to predict the annotation scores, similar to a regression task.

\subsection{Evaluation and Discussion}
Partitioning our corpus as 80\% training, 10\% development, and 10\% test sets, we evaluated our model using two measures: (i) total root mean square error (RMS) - the deviation of the model predictions from the human-annotations and, (ii) $F_1$ measure in a classification task - the ability of the model to predict human-annotations categorized into a group of classes. To evaluate performance in terms of classification, we discretized annotation scores into 6 classes: [0, 0.1), [0.1, 0.3), [0.3, 0.5), [0.5, 0.7), [0.7, 0.9), [0.9, 1].

Table \ref{model-performance} summarizes the performance of our models on the test set, presenting average scores for 5 different configurations, to compensate for outlier results due to randomness in model initialization. While the LSTM-CRF had a better (higher) F-score on the classification task, its RMS score was worse (higher) than the LSTM-SIG model, which may be due to the limitation of the model as discussed in Section 5.

\begin{table}[!ht]
\centering
\begin{tabular}{@{}lll@{}}
\toprule
\textbf{Model} & \textbf{RMS} & \textbf{$F_1$ (macro)} \\ \midrule
LSTM-CRF       &        0.154             &  \textbf{0.60}                     \\
LSTM-SIG       &       \textbf{0.120}             &  0.519                      \\ \bottomrule
\end{tabular}
\caption{Model performance in terms of RMS deviation and macro-averaged $F_1$ score, with best results in \textbf{bold} font.}
\label{model-performance}
\end{table}

Confusion matrices in Figure \ref{conf-mat} provide a more detailed view of the classification performance of each model. Since the LSTM-SIG was trained to optimize the accuracy of its continuous predictions, rather than its discrete assignment of instances to classes, it is not surprising to see a ``wider diagonal'' in the confusion matrix in Figure \ref{conf-mat}(b), which indicates that the LSTM-SIG model was more likely to misclassify words using ordinally adjacent classes. The figure illustrates that both models were worse at classifying words with importance scores in the middle range [0.3, 0.7).


Treating our human-annotations as ground truth, we also computed the concordance correlation coefficient to measure the agreement between the human annotation and each model. The average correlation between the human annotator and the LSTM-CRF model was higher ($\rho_c = 0.839$), as compared to the LSTM-SIG model ($\rho_c = 0.826$).

\section{Conclusions and Future Work}

We have presented a new collection of annotation of transcripts of the Switchboard conversational speech corpus, produced through human annotation of the importance of individual words to the meaning of each utterance. We have demonstrated the use of this data by training word-importance prediction models, with the best model achieving an $F_1$ score of $0.60$ and model-human agreement correlation of $0.839$. In future work, we will collect additional human annotations for additional sections of the corpus. This research is part of a project on the use of ASR to provide real-time captions of speech for DHH individuals during meetings, and we plan to incorporate these word-importance models into new word-importance-weighted metrics of ASR accuracy, to better predict the usability of ASR-produced captions for these users.

\section{Acknowledgement}
This material was based on work supported by the National Technical Institute for the Deaf (NTID). We thank Tomomi Takeuchi and Michael Berezny for their contributions.

\section{Bibliographical References}
\label{main:ref}

\bibliographystyle{lrec}
\bibliography{main}

\end{document}